%% file: main.tex
\documentclass[11pt,a4paper]{article}
\usepackage[hyperref]{eacl2021}
\usepackage{times}
\usepackage{latexsym}
\usepackage{float}

\usepackage{subfiles}   
\usepackage{graphicx}
\usepackage{amsmath}
\usepackage{amsfonts}
\usepackage{microtype}
\usepackage{enumitem}
\setlist{nosep}

\aclfinalcopy 
\def\aclpaperid{1074} 

\newcommand{\cut}[1]{}
\usepackage{xcolor}
\newcommand{\xhdr}[1]{\noindent{\bfseries #1}.}
\newcommand{\CITE}{\textcolor{red}{CITE }}

\title{Exploring the Limits of Few-Shot Link Prediction in Knowledge Graphs}

\author{Dora Jambor\textsuperscript{1,2},
  Komal Teru\textsuperscript{1,2},
  Joelle Pineau\textsuperscript{1,2,3},
  William L. Hamilton\textsuperscript{1,2} \\
  \textsuperscript{1} School of Computer Science, McGill University, Canada \\
  \textsuperscript{2} Quebec AI Institute (Mila), Canada \\
  \textsuperscript{3} Facebook AI Research (FAIR)\\
  \{dora.jambor, komal.teru, jpineau, wlh\}\\@\{mail.mcgill.ca, mail.mcgill.ca, cs.mcgill.ca, cs.mcgill.ca\}}

\date{}

\begin{document}
\maketitle
\begin{abstract}
Real-world knowledge graphs are often characterized by low-frequency relations---a challenge that has prompted an increasing interest in few-shot link prediction methods. 
These methods perform link prediction for a set of new relations, unseen during training, given only a few example facts of each relation at test time. 
In this work, we perform a systematic study on a spectrum of models derived by generalizing the current state of the art for few-shot link prediction, with the goal of probing the limits of learning in this few-shot setting. 
We find that a simple {\em zero-shot} baseline---which ignores any relation-specific information---achieves surprisingly strong performance. Moreover, experiments on carefully crafted synthetic datasets show that having only a few examples of a relation fundamentally limits models from using fine-grained structural information and only allows for exploiting the coarse-grained positional information of entities. Together, our findings challenge the implicit assumptions and inductive biases of prior work and highlight new directions for research in this area.

\end{abstract}

\section{Introduction}

\subfile{sections/introduction}

\section{Few-shot link prediction}

\subfile{sections/methodology}

\section{Null Models}
\label{sec:null_models}

\subfile{sections/nullmodels}

\section{Experiments}

\subfile{sections/experiments}

\section{Conclusion}

\subfile{sections/conclusion}

\section*{Acknowledgments}
The authors would like to thank Priyesh Vijayan, Joey Bose, Lu Liu and other members of the RLLab and Mila for their invaluable feedback and useful discussions on the early drafts. Furthermore,  the authors are grateful for the anonymous reviewers for their comments on the first draft of the paper.
This research was supported in part by Canada CIFAR AI Chairs held by Prof. Pineau and Prof. Hamilton, as well as gift grants from Microsoft Research and Samsung AI.


\bibliography{anthology,eacl2021}
\bibliographystyle{acl_natbib}

\clearpage
\newpage

\appendix

\section{Hyperparameters}
\label{sec:appendix_a}
As discussed, we followed the experimental setup described in \citet{chen2019meta}. We used the Adam optimizer \citep{kingma2014adam} with a learning rate of 0.001, using 1 to 3 ratio of positive to negative samples. During training, we used 3 queries per task on each dataset.

We adapted the batch size to be 1024, and the number of queries to test on to be 3, based on their open-sourced codebase. These hyperparameters yielded the best performing models.
Similarly, we also used the same train/validation/test relation splits of 51:5:11 and 133:16:34 for Nell-One and Wiki-One respectively.

For our R-GCN model, we considered a range of [5, 10, 20] as the number of neighbors to sample for each message passing step, and [2, 4] as the number of basis. Furthermore, we used 2 layers in the R-GCN.
Finally, we used 50 and 20 as the R-GCN hidden layer dimension for Nell-One and Wiki-One, respectively. These hyperparameters were partly followed from \citet{schlichtkrull2018modeling}, and were decided upon consideration for our available compute infrastructure.

Our models were trained on a single Nvidia 1080Ti GPU, and each model training took between 13-18 hours depending on the model and dataset settings.

\begin{figure}[b!]
\begin{center}
    \includegraphics[width=\textwidth]{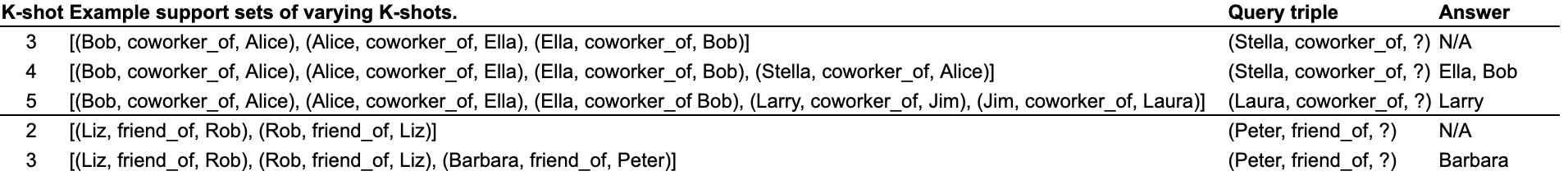}
    \caption{Limits of logical inference in the few-shot domain.}
    \label{fig:limits}
\end{center}
\end{figure}

\begin{figure}[tbt!]
\centering
    \includegraphics[width=\textwidth]{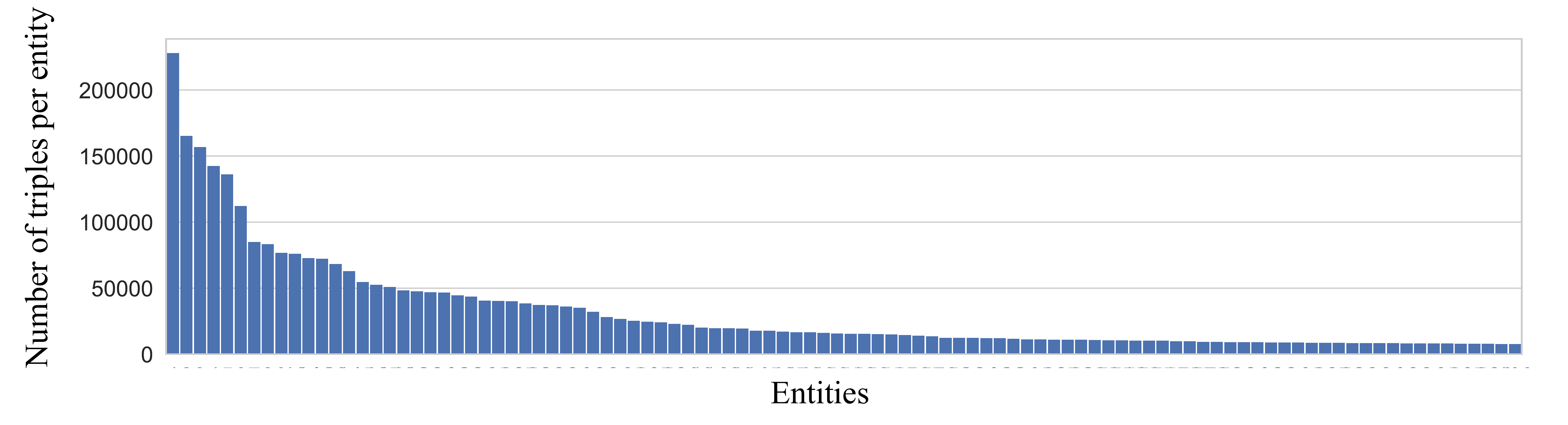}
    \caption{Entity degree distribution of Wiki-One}
    \label{fig:wiki_degree}
\end{figure}

\section{Logical inference in the few-shot domain}
\label{sec:appendix_b}

Figure \ref{fig:limits} shows five example support sets to demonstrate that inferring logical properties such as symmetry and transitivity requires a minimum number of carefully designed K-shot examples. 

\section{Entity Frequency Analysis}
\label{sec:ent_frequency}

Figure \ref{fig:wiki_degree} shows the 100 highest-degree entities out of all 4,838,244 entities in the Wiki-One knowledge graph. We find that the median degree of entities is 1, and the highest degree is 227,390, which connects to 4.69\% of the total graph. We suspect that these high-degree entities, so-called hub nodes, may add noise to the embeddings of support set entities. This could in turn affect performance and explain why we do not observe a strong correlation between the degrees of support set entities and performance in Wiki-One.

\end{document}

%% file: sections/introduction.tex
A knowledge graph (KG) is a multi-relational graph that offers a structured way to organize facts about the world.
Encoder-decoder approaches are commonly used to predict new facts from existing ones where entities and relations are embedded in a low-dimensional vector space via an encoder, to then score the likelihood of observing a new fact via a decoder \citep{nickel2015review, transE, complex, dettmers2018conve}.

It is well known that the performance of these methods can significantly drop when predicting for relations that are only observed in a few example facts. However, link prediction for these low-frequency relations is very important, as not only are these relations abundant in most knowledge graphs, they are also key for knowledge graph completion tasks where new relations may appear after model training.

To study this low-frequency regime, \citet{xiong2018one} created the Nell-One and Wiki-One benchmarks where the task is to predict new facts for a set of new relations at test time, where each relation is only observed a \textit{few} times (as specified by some small fixed number $K$). 
Previous approaches have shown promising results using metric-based \citep{xiong2018one} and gradient based meta-learning techniques \citep{chen2019meta}.
However, we argue that these models are limited by the current task formulation to only exploit coarse-grained positional signals (i.e., nodes belonging to the same community) that are abundant in these benchmarks, rather than leveraging structural signals (e.g. transitivity, symmetry).


\begin{figure}[t]
    \centering
    \includegraphics[width=\linewidth]{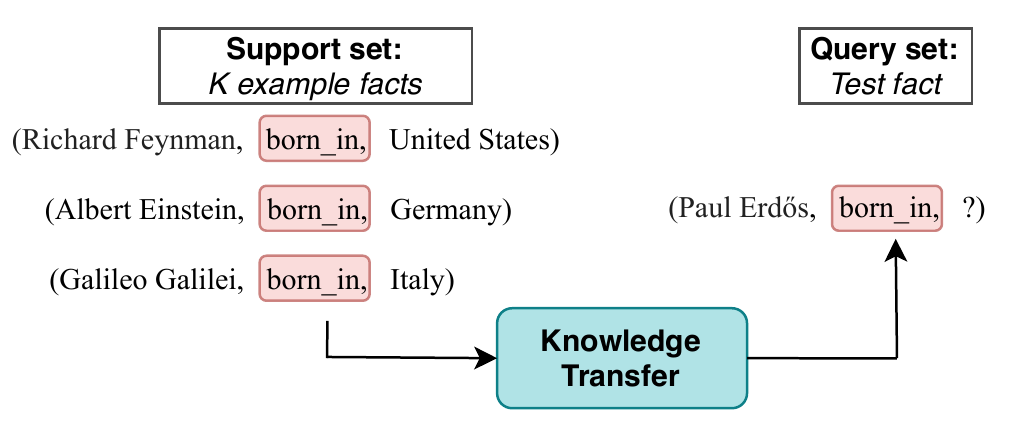}
    \vspace{-0.3in}
    \caption{The few-shot link prediction task}
    \label{fig:task}
    \vspace{-0.25in}
\end{figure}

\xhdr{Present work}
In this work, we take a critical take on current approaches for few-shot link prediction over knowledge graphs. We posit that current meta-learning based approaches benefit largely due to the positional signals in entities, rather than utilising information about the low frequency relations. We corroborate these insights by conducting a systematic study on a spectrum of models with decreasing complexity.    
Interestingly, we find that a much simpler zero-shot variant of the state of the art ---devoid of any meta-learning scheme--- yields surprisingly competitive results, while not consuming any example facts about a relation. 
Motivated by these observations, we design a set of null models tailored to different learning signals a model might utilize to drive effective link prediction. Empirically, we validate that these existing meta-learning models are ill-equipped to infer logical patterns about the few-shot relations. These findings bring forth the shortcomings of the current task formulation and raises new questions in both task and model design while highlighting new directions of research in few-shot link prediction.

%% file: sections/methodology.tex
\subsection{Problem definition}
The goal of few-shot link prediction is to predict missing links for a new relation by only observing  $K$ example triples of that relation (Figure \ref{fig:task}). Following literature in few-shot classification \citep{ravi2016optimization, vinyals2016matching, snell2017prototypical}, we organize our dataset as a set of tasks, where a task corresponds to predicting links for a new relation. The set of tasks for training and testing are disjoint, with the added constraint that entities in the test tasks are a subset of the entities in the train tasks.
Let $\mathcal{V}$ denote the set of entities in the knowledge graph. For each new relation $r_i$, we then construct a support set $\mathcal{S}_{i} = \{(h_k, r_i, t_k)\}_{k=1}^K$ containing $K$ example entity pairs, $h_k, t_k \in \mathcal{V}$, connected by relation $r_i$, and a query set $\mathcal{Q}_{i} = \{(h_j, r_i, ?)\}_{j=1}^J$ containing $J$ query triples over entities in $\mathcal{V}$. As shown in Figure \ref{fig:task}, the goal is then to learn how to extract knowledge from the support set such that we can predict the missing tail entities in the query set.




\subsection{Overview of the framework}
The foundation of our analyses focuses on a generalization of the current state-of-the-art gradient-based meta-learning approach \citep{chen2019meta}.
This approach follows the encoder-decoder paradigm of embedding-based knowledge graph completion methods \citep{hamilton2017representation}, where the entities and relations are embedded in a low-dimensional vector space and the embeddings are used to predict the likelihood of a given triple.

\xhdr{Encoder functions} The key idea in few-shot learning is to transfer knowledge from support set to query set by learning a function $\texttt{RelLearner}: \mathcal{S}_i \mapsto \mathbb{R}^d$. This maps a support set $\mathcal{S}_{i}$, which characterizes the relation $r_i$, to a low dimensional embedding via an encoder function $\texttt{E}: \mathcal{V} \mapsto \mathbb{R}^d$
\begin{equation}
    \mathbf{r}_{i} = \texttt{RelLearner}(\{(\texttt{E}(h_k), \texttt{E}(t_k)\}_{k=1}^K).
\label{eq:rel_learner}
\end{equation}
The \texttt{RelLearner} function can vary from a simple MLP \citep{hastie2009elements} to more complicated recurrent architectures \citep{rumelhart1985learning, jordan1997serial, hochreiter1997long}. Further, the entity encoder \texttt{E} can vary from TransE-style embeddings \cite{transE, sun2018rotate} to a graph neural network \cite{schlichtkrull2018modeling} that explicitly leverages the neighborhood information around entities. 

\xhdr{Decoder and loss function} A decoder function ingests the embeddings of the entities $\mathbf{h}, \mathbf{t}$ and of the relation $\mathbf{r}$ to score the likelihood of a given triple $(h, r, t)$. 
Using a simple TransE decoder \cite{transE}, it is then optimized to score positive triples higher than negative triples using a contrastive loss $\mathcal{L}$ \citep{dyer2014notes}. In the few-shot setting we compute the support set loss $\mathcal{L}(\mathcal{S}_i)$, and the final query set loss $\mathcal{L}(\mathcal{Q}_i)$, which are used to update the model parameters \cite{chen2019meta}.

\xhdr{Meta-gradient update}
Instead of directly using the relation embedding $\mathbf{r}_i$ from Equation \eqref{eq:rel_learner} to compute the final query loss $\mathcal{L}(\mathcal{Q}_i)$, we first make an update on the relation embedding using the gradient of the support set loss $\mathcal{L}(\mathcal{S}_i)$
\begin{equation}
    \mathbf{r}_i' = \mathbf{r}_i - \eta \nabla_{\mathbf{r}_i} \mathcal{L}(\mathcal{S})
\label{eq:metagrad}
\end{equation}
where $\eta$ denotes the learning rate.
This update encourages $\mathbf{r}_i$ to be such that it effectively predicts the support set triples via minimizing $\mathcal{L}(\mathcal{S}_i)$. 

\begin{table*}[h!]
\centering
\begin{tabular}{llllllllll}
\hline
 &&
  \multicolumn{2}{c}{MRR} &
  \multicolumn{2}{c}{Hits@10} &
  \multicolumn{2}{c}{Hits@5} &
  \multicolumn{2}{c}{Hits@1} \\
 &&
  \multicolumn{1}{c}{1-shot} &
  \multicolumn{1}{c}{5-shot} &
  \multicolumn{1}{c}{1-shot} &
  \multicolumn{1}{c}{5-shot} &
  \multicolumn{1}{c}{1-shot} &
  \multicolumn{1}{c}{5-shot} &
  \multicolumn{1}{c}{1-shot} &
  \multicolumn{1}{c}{5-shot} \\
\hline
\textbf{Nell-One} &
MetaR & 
  0.294 &
  \multicolumn{1}{l}{0.323} &
  0.464 &
  \multicolumn{1}{l}{0.500} &
  0.398 &
  \multicolumn{1}{l}{0.426} &
  0.201 &
  0.230 \\
&SharedEmbed & 
  0.276 &
  \multicolumn{1}{l}{0.311} &
  0.454 &
  \multicolumn{1}{l}{0.495} &
  0.382 &
  \multicolumn{1}{l}{0.420} &
  0.173 &
  0.205 \\
&ZeroShot &
  0.199 &
  \multicolumn{1}{l}{0.219} &
  0.342 &
  \multicolumn{1}{l}{0.365} &
  0.283 &
  \multicolumn{1}{l}{0.303} &
  0.116 &
  0.136 \\
&R-GCN &
  0.216 &
  \multicolumn{1}{l}{0.267} &
  0.412 &
  \multicolumn{1}{l}{0.464} &
  0.316 &
  \multicolumn{1}{l}{0.366} &
  0.120 &
  0.172 \\
\hline
{\textbf{Wiki-One}*} &
MetaR & 
  0.325 &
  \multicolumn{1}{l}{0.326} &
  0.448 &
  \multicolumn{1}{l}{0.408} &
  0.408 &
  \multicolumn{1}{l}{0.367} &
  0.258 &
  0.280 \\
&SharedEmbed &
  0.290 &
  \multicolumn{1}{l}{0.311} &
  0.399 &
  \multicolumn{1}{l}{0.415} &
  0.348 &
  \multicolumn{1}{l}{0.378} &
  0.238 &
  0.254 \\
&ZeroShot &
  0.279 &
  \multicolumn{1}{l}{0.289} &
  0.361 &
  \multicolumn{1}{l}{0.367} &
  0.337 &
  \multicolumn{1}{l}{0.341} &
  0.234 &
  0.246 \\
&R-GCN &
  0.126 &
  \multicolumn{1}{l}{0.137} &
  0.178 &
  \multicolumn{1}{l}{0.237} &
  0.130 &
  \multicolumn{1}{l}{0.152} &
  0.101 &
  0.104 \\
  \hline
\end{tabular}
\vspace{-5pt}
\caption{\label{tab:test_results}
Average metrics on Nell-One and Wiki-One few-shot link prediction tasks. * For Wiki-One we used pre-trained embeddings, similar to \citet{chen2019meta}.}
\vspace{-5mm}
\end{table*}

\subsection{Baselines}
 \label{sec:models}
 Our objective is to probe how much models leverage the support set to perform the query task.
 To this end, we perform a systematic study of different model variants, where each falls into the general framework described in Section 2.2.


\textbf{MetaR} follows \citet{chen2019meta}, where the \texttt{RelLearner} is defined as a 2-layer MLP \citep{hastie2009elements} over the support set entity embeddings. The encoder \texttt{E} simply maps each entity to a fixed learnable vector as in \citet{transE}.

\textbf{SharedEmbed} skips Equation \eqref{eq:rel_learner}, and instead sets $\mathbf{r_i} = \mathbf{r_g}$, where $\mathbf{r_g}$ is a single learnable embedding shared across all relations.
We propose this modification to measure the effect of representing all relations by the same embedding $\mathbf{r_g}$, where the only information from the support set comes via the gradient update in Equation \eqref{eq:metagrad}.

\textbf{ZeroShot} further removes the meta-gradient update in Equation \eqref{eq:metagrad} and lets $\mathbf{r_i'} = \mathbf{r_g}$. This effectively reduces the model to perform zero-shot link prediction on the relation's query set without any relation-specific information from the support set.

\textbf{R-GCN} uses the same \texttt{RelLearner} as in MetaR, with the exception that support set entity embeddings are learned via a multi-relational graph neural network, R-GCN \citep{schlichtkrull2018modeling}, instead of a TransE-style embeddings. R-GCN learns entity representation via aggregating the 2-hop neighbors of a given entity. With this model we probe the extent to which injecting structural bias into entity representations can influence performance in link prediction.



%% file: sections/nullmodels.tex
\begin{figure*}[t]
    \centering
    \includegraphics[width=\textwidth]{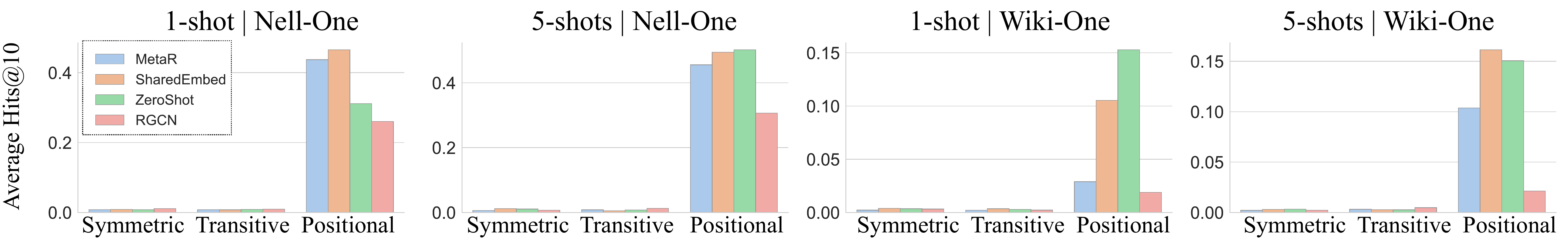}
    \caption{Average Hits@10 on synthetically generated relations using our proposed null models.}
    \label{fig:synthetic_results}
    \vspace{-2mm}
\end{figure*}

In order to probe and understand the performance of different models, we introduce two \textit{null models}, which are used to generate synthetic data that satisfy certain properties.
Motivated by recent literature on position versus structure-aware methods in relational learning \citep{you2019position, Srinivasan2020On},
we test the models' ability to learn from two key sources of information: \textit{structural information} and \textit{positional information}.
In the context of knowledge graphs, \textit{structural information} corresponds to the fine-grained relational semantics. These are the logical patterns that are extracted by state-of-the-art rule induction systems, such as RuleN \citep{meilicke2018fine}. 

On the other hand, \textit{positional information} corresponds to the coarse-grained community structure of the nodes in the graph. In other words, two nodes are said to be positionally `close' in the graph, i.e., if they belong to the same community \cite{newman2018networks}.



\subsection{Structural Null Models} 

The first type of null models contains synthetic relations that satisfy simple logical properties.
For the sake of exposition, we focus on two simple logical patterns: symmetry and transitivity.
For the purposes of all synthetic data generation, we only consider the largest connected component of respective datasets, denoted $\mathcal{G}^L$.

\xhdr{Synthetic symmetric relations} To generate $2N$ edges connected by a symmetric relation $r^*_\text{s}$, we repeat the following steps $N$ times.:
\begin{enumerate}[nosep]
    \item Uniformly sample a pair of unique entities--$h_i, t_i$--from all the entities in $\mathcal{G}^L$.
    \item Add two edges--${((h_i, r^*_\text{s}, t_i)), ((t_i, r^*_\text{s}, r_i))}$ to the set of synthetic symmetric edges.
\end{enumerate}

\xhdr{Synthetic transitive relations} To sample $3N$ edges connected by a transitive relation $r^*_\text{t}$, we generate 3 edges at a time. In particular, we repeat the following steps $N$ times:
\begin{enumerate}[nosep]
    \item Uniformly sample 3 unique entities--$e_1$, $e_2$, and $e_3$--from all the entities in $\mathcal{G}^L$.
    \item Add three edges--$(e_1, r^*_\text{t}, e_2)$, $(e_2, r^*_\text{t}, e_3)$, $(e_1, r^*_\text{t}, e_3)$--to our collection.
\end{enumerate}

\subsection{Positional Null Models}
The second type of null models focuses on generating synthetic relations that depend on the underlying community structure in the graph.
We call these relations {\em positional} because they depend on the relative global position of the entities, rather than on local structural properties.

We first cluster the largest connected component $\mathcal{G}^L$ into $K$ communities using a standard algorithm originally proposed by \citet{blondel2008fast}. Let $\{C_i\}_{i=1}^K$ denote the set of communities generated, where each community is a set of entities from $\mathcal{G}^L$. To generate $N$ synthetic edges for a positional relation $r_p$, we repeat the following steps $N$ times:
\begin{enumerate}[nosep]
    \item Uniformly sample a community index $i$ from the set $\{1,,K\}$.
    \item Uniformly sample two unique entities $h, t$ from community $C_i$; add $(h,r^*_\text{p},t)$ to the set.
\end{enumerate}

\cut{
The second type of null models focuses on generating synthetic relations that depend on the underlying community structure in the graph.
We call these relations {\em positional} because they tend to connect entities that come from the same communities in the graph. These relations depend on the relative global position of the entities, rather than on local structural properties.

The general strategy for generating random triples for a positional synthetic relation $r^*$ is summarized as follows:
\begin{enumerate}
    \item Let $G^L$ denote the largest component of the knowledge graph. We again define $E$ as the set of entities that belongs to $G^L$.

    \item Find K communities of $G^L$ using the community detection approach described in \citet{blondel2008fast}. 
    
    \item Let $F=\emptyset$ denote the set of synthetic triples for $r^*$. Repeat the following steps for $N$ iterations:
    \begin{enumerate}
        \item Uniformly sample a community index $i$ from the set $\{1,...,K\}$.
        \item Uniformly sample two entities $h, t$ without replacement from community $C_i$.
        \item Add $(h,r^*,t)$ to the set $F$.
    \end{enumerate}
\end{enumerate}}


%% file: sections/experiments.tex
We followed the same experimental setup as in \citet{chen2019meta}, as described in Appendix \ref{sec:appendix_a}.
We conducted our experiments on the Nell-One and Wiki-One datasets\footnote{Datasets can be downloaded under \href{https://www.dropbox.com/sh/d04wbxx8g97g1rb/AABDZc-2pagoGhKzNvw0bG07a?dl=0}{this link}.}. For more details on these benchmarks, we refer the reader to Table 1 in \citet{xiong2018one}. Similar to earlier work, we report MRR, Hits@1, Hits@5 and Hits@10 on our test relations, using a type-constrained candidate set.


\subsection{Results and Analysis}
\begin{figure}[t]
    \centering
    \includegraphics[width=\columnwidth]{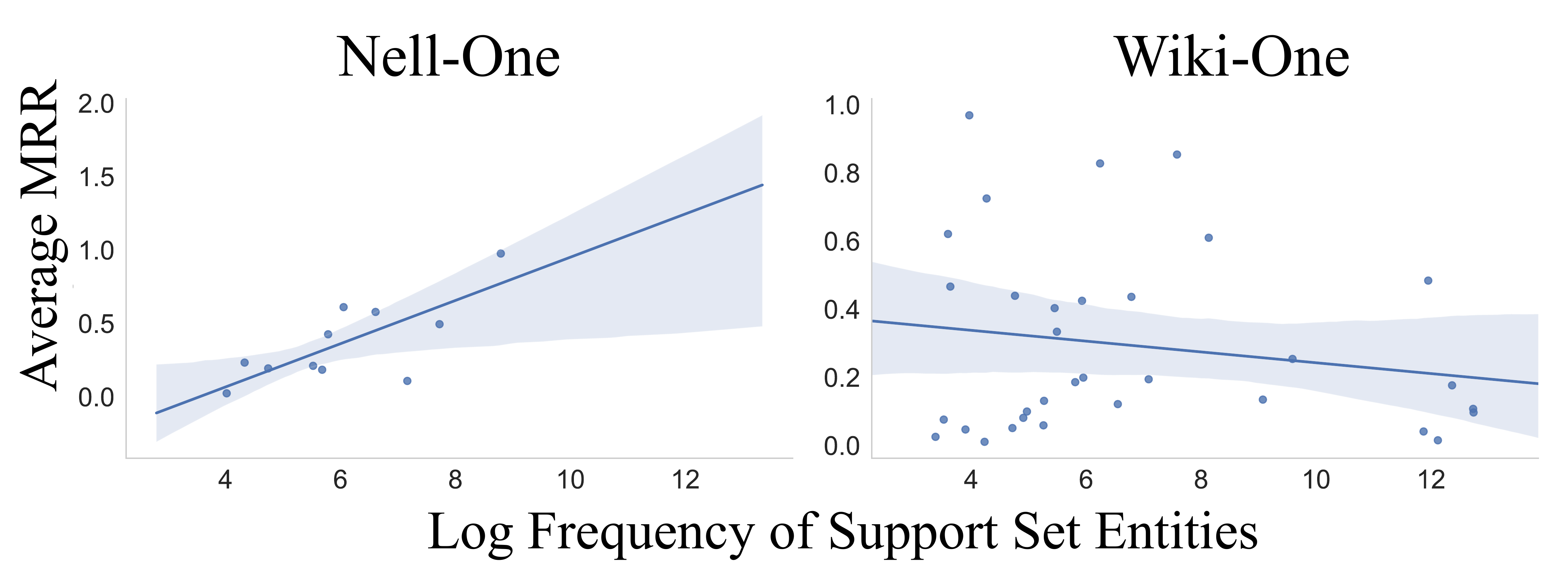}
    \caption{Pearson's R between MRR and the log frequency of support set entities in training graph. 
    We observe strong correlation for Nell-One, but not for Wiki-One. For more details on this, see Appendix \ref{sec:ent_frequency}.
    }
    \label{fig:correlation}
\vspace{-4mm}
\end{figure}

\xhdr{Experiments on Real Data}
As shown in Table \ref{tab:test_results}, for Nell-One, we find that SharedEmbed model yields competitive performance to MetaR, with Hits@10 of 45.4\% and 49.5\%, as compared to MetaR's Hits@10 of 46.4\% and 50.0\%  for 1 and 5-shot, respectively. 
The same observation holds for Wiki-One, where SharedEmbed yields 39.9\% and 41.5\% Hits@10, compared to MetaR's Hits@10 of 44.8\% and 40.8\%, for 1 and 5-shot, respectively. 

It is surprising how competitive SharedEmbed is, given that the only relation-specific information the model gets to observe comes via the meta-gradient update in Equation \eqref{eq:metagrad}. 
In fact, we find that even in absence of this gradient signal, i.e., without \textit{any} relation-specific information, ZeroShot performs relatively good, with Hits@10 of 34.2\% and 36.5\% on Nell-One, and 36.1\% and 36.7\% on Wiki-One.

The nontrivial performance of these simple models suggests that such models may exploit some easily accessible positional signals around entities, without the need to learn meaningful representations for relations. 
In fact, Figure \ref{fig:correlation} shows a high correlation between performance and the degrees of entities in the support set for Nell-One. 
We reconcile this observation by noting that as models observe more signals about entities, they start relying less on the support set, and thus on the relation representations. 
Furthermore, contrary to our expectation, even when we equip models with structural biases, as done via an R-GCN, they do not yield better results.

\xhdr{Null Model Experiments}
We probed the above trained models on the synthetically generated test tasks following the procedure discussed in Section \ref{sec:null_models}.
As shown in Figure \ref{fig:synthetic_results}, we find a consistent trend for these models to yield higher performance on tasks that rely on positional signals, as compared to tasks that require logical inference.

Indeed, in the current task formulation, where we are given a support set of K randomly sampled examples, it is unlikely that logically consistent patterns will be captured in the K-shot examples. 
For example, seeing conclusive evidence of transitivity when only given a small random sample of tuples is highly unlikely. 
In fact, as we show in Appendix \ref{sec:appendix_b}, one provably cannot learn certain logical patterns for some values of K in the K-shot setting.



%% file: sections/conclusion.tex
We conducted a systematic study of various models to probe their limits in performing few-shot link prediction. 
Our experiments on both synthetic and real data show that the current task formulation encourages models to mainly rely on positional information around entities, rather than leveraging logical signals about relations.
In fact, we empirically show that having only K examples of a relation fundamentally limits the types of logical patterns that can be learned.
We argue that a future direction in few-shot link prediction should allow for a more careful construction of the support set, to scaffold the use of logical patterns in few-shot learning.